\pdfoutput=1

\documentclass[11pt]{article}

\usepackage{acl}

\usepackage{times}
\usepackage{latexsym}

\usepackage[T1]{fontenc}

\usepackage[utf8]{inputenc}

\usepackage{microtype}

\usepackage{amsmath,amsthm,amsfonts,amssymb,amscd}
\usepackage{mathrsfs}
\usepackage{times}
\usepackage{graphicx} 
\usepackage{wrapfig}
\usepackage{subcaption}
\usepackage{multirow}
\usepackage{arydshln}
\usepackage{booktabs}
\usepackage{array}
\usepackage{rotating}
\usepackage{pgfplots} 

\title{Model-based Subsampling for Knowledge Graph Completion}

\author{
  Xincan Feng\textsuperscript{\dag}, Hidetaka Kamigaito\textsuperscript{\dag}, Katsuhiko Hayashi\textsuperscript{\ddag}, Taro Watanabe\textsuperscript{\dag} \\
  \textsuperscript{\dag}Nara Institute of Science and Technology \textsuperscript{\ddag}Hokkaido University \\
  \texttt{\{feng.xincan.fy2, kamigaito.h, taro\}@is.naist.jp} \\
  \texttt{katsuhiko-h@ist.hokudai.ac.jp}}

\begin{document}
\maketitle
\begin{abstract}
Subsampling is effective in Knowledge Graph Embedding (KGE) for reducing overfitting caused by the sparsity in Knowledge Graph (KG) datasets. However, current subsampling approaches consider only frequencies of queries that consist of entities and their relations. Thus, the existing subsampling potentially underestimates the appearance probabilities of infrequent queries even if the frequencies of their entities or relations are high. 
To address this problem, we propose Model-based Subsampling (MBS) and Mixed Subsampling (MIX) to estimate their appearance probabilities through predictions of KGE models. Evaluation results on datasets FB15k-237, WN18RR, and YAGO3-10 showed that our proposed subsampling methods actually improved the KG completion performances for popular KGE models, RotatE, TransE, HAKE, ComplEx, and DistMult.
\end{abstract}

\section{Introduction}
\label{sec:intro}

A Knowledge Graph (KG) is a graph that contains entities and their relations as links. KGs are important resources for various NLP tasks, such as dialogue \cite{moon-etal-2019-opendialkg}, question-answering \cite{10.1145/3038912.3052675}, and natural language generation \cite{Guan_Wang_Huang_2019}, etc.
However, covering all relations of entities in a KG by humans takes a lot of costs. Knowledge Graph Completion (KGC) tries to solve this problem by automatically completing lacking relations based on the observed ones.
Letting $e_i$ and $e_k$ be entities, and $r_j$ be their relation, KGC models predict the existence of a link $(e_i,r_j,e_k)$ by filling the $?$ in the possible links $(e_i,r_j,?)$ and $(?,r_j,e_k)$, where $(e_i,r_j)$ and $(r_j,e_k)$ are called queries, and the $?$ are the corresponding answers.

\begin{figure}[t]
    \centering
    \begin{tikzpicture}
    \begin{axis} [
    xlabel = Mean Reciprocal Rank (MRR),
    xlabel style={font=\small},
    x tick label style={font=\small},
    yticklabels = {X,YAGO3-10,WN18RR,FB15k-237},
    y tick label style={
        font=\small,
        rotate=90},
    xbar = .05cm,
    bar width = 12pt,
    xmin = 25, 
    xmax = 55, 
    enlarge y limits = {abs = .7},
    legend style={
    font=\small,
    at={(1,1)},
    anchor=north east,
    }
    ]
    \addplot coordinates {(51.8,0) (42.0,1) (32.8,2)}; 
    \addplot coordinates {(51.4,0) (41.2,1) (28.6,2)}; 
    \legend {w/ Subsampling, w/o Subsampling};
    \end{axis}
    \end{tikzpicture}
    \caption{The averaged KGC performance (MRR) of KGE models\footnotemark with and without subsampling on FB15k-237, WN18RR, and YAGO3-10.}
    \label{fig:effect_subsampling}
\end{figure}
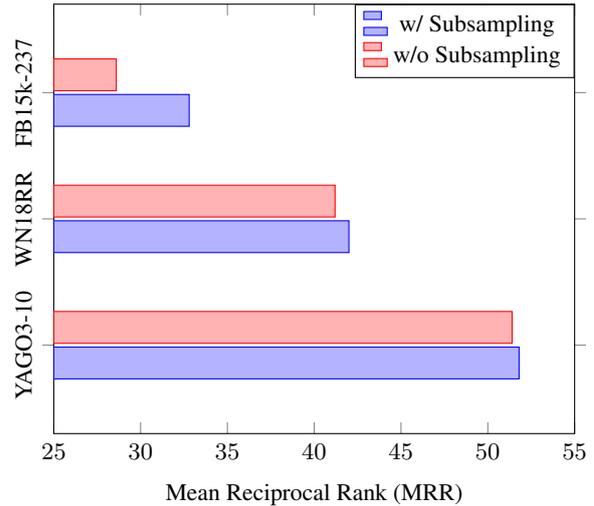

Currently, Knowledge Graph Embedding (KGE) is a dominant approach for KGC.
KGE models represent entities and their relations as continuous vectors. Since the number of these vectors proportionally increases to the number of links in a KG, KGE commonly relies on Negative Sampling (NS) to reduce the computational cost in training.
In NS, a KGE model learns a KG by discriminating between true links and false links created by sampling links in the KG. While NS can reduce the computational cost, it has the problem that the sampled links also reflect the bias of the original KG.

\footnotetext{See Appendix \ref{appendix:settings} for the details.}

As a solution, \newcite{rotate} introduce subsampling \cite{ns} into NS for KGE. In this usage, subsampling is a method of mitigating bias in a KG by discounting the appearance frequencies of links with high-frequent queries and reserving the appearance frequencies for links with low-frequent queries.
Figure \ref{fig:effect_subsampling} shows the effectiveness of using subsampling. From this figure, we can understand that KGE models cannot perform well without subsampling on commonly used datasets such as FB15k-237~\cite{Tou15}, WN18RR~\cite{Det18}, and YAGO3-10~\cite{Det18}. Furthermore, the improved MRR on FB15k-237, which has more sparse relations than the other datasets, indicates that subsampling actually works on the sparse dataset.

\begin{figure}[t]
    \centering
    \includegraphics[width=\columnwidth]{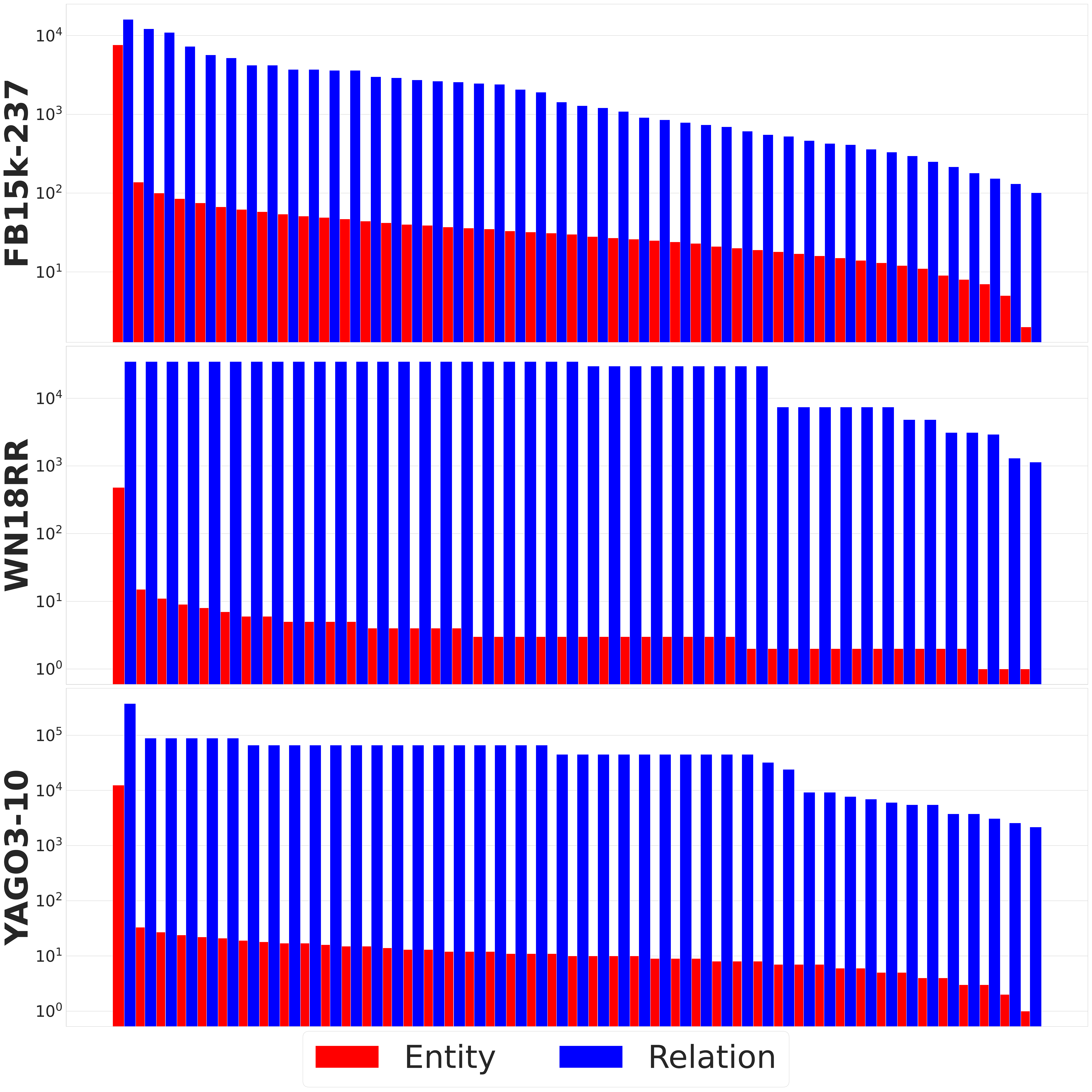}
    \caption{Frequencies of entities and relations included in each query that appeared only once in training data of FB15k-237, WN18RR, and YAGO3-10\footnotemark.}
    \label{fig:e_and_r_freqs}
\end{figure}

However, the current subsampling approaches in KGE \cite{rotate,pmlr-v162-kamigaito22a} only consider the frequencies of queries.
Thus, these approaches potentially underestimate the appearance probabilities of infrequent queries when the frequencies of their entities or relations are high.
Figure \ref{fig:e_and_r_freqs} shows the frequencies of entities and relations included in each query that appeared only once in training data.
From the statistics, we can find that the current count-based subsampling (CBS) does not effectively use frequencies of entities and relations in infrequent queries, although these have sufficient frequencies.
\footnotetext{Due to the space limitation, it is difficult to plot all the values in this graph. Thus, we filter the entities and relations for every certain amount after they are sorted by frequency in descending order. The filtering amounts for FB15k-237, WN18RR, and YAGO3-10 are 2,000, 1,778, and 4,444, respectively.
By this filtering, the number of plotted entities and relations for FB15k-237, WN18RR, and YAGO3-10 are reduced to 45, 44, and 45, respectively.}

To deal with this problem, we propose Model-based Subsampling (MBS) that can handle such infrequent queries by estimating their appearance probabilities through predictions from KGE models in subsampling. 
Since the observed frequency in training data does not restrict the estimated frequencies of MBS different from CBS, we can expect the improvement of KGC performance using MBS. 
In addition, we also propose Mixed Subsampling (MIX), which uses the frequencies of both CBS and MBS to boost their advantage by reducing their disadvantages. 

In our evaluation on FB15k-237, WN18RR, and YAGO3-10 datasets, we adopted our MBS and MIX to the popularly used KGE models RotatE~\cite{rotate}, TransE~\cite{Bor13}, HAKE~\cite{https://doi.org/10.48550/arxiv.1911.09419}, ComplEx~\cite{Tro16}, and DistMult~\cite{yang2015embedding}. 
The evaluation results showed that MBS and MIX improved MRR, H@1, H@3, and H@10 from Count-based Subsampling (CBS) in each setting\footnote{Our code is available on \url{https://github.com/xincanfeng/ms_kge}.}.

\section{Subsampling in KGE}

\subsection{Problem Definitions and Notations}
\label{subsec:notations}

We denote a link of a KG in the triplet format $(h,r,t)$. $h$ is the head entity, $t$ is the tail entity, and $r$ is the relation of the head and tail entity. In a classic KG completion task, we input the query $(h,r,?)$ or $(?,r,t)$, and output the predicted head or tail entity corresponding to $?$ as the answer. More formally, let us denote the input query as $x$ and its answer as $y$, hereafter. A score function $s_{\mathbf{\theta}}(x,y)$ predicts $p_{\mathbf{\theta}}(y|x)$, a probability for a given query $x$ linked to an answer $y$ based on a model $\mathbf{\theta}$. In general, we train $\mathbf{\theta}$ by predicting $p_{\mathbf{\theta}}(y|x)$ on $|D|$ number of links, where $D=\{(x_{1},y_{1}),\cdots, (x_{|D|},y_{|D|})\}$ is a set of observables that follow $p_d(x, y)$.

\subsection{Negative Sampling in KGE}

Since calculating all possible $y$ for given $x$ is computationally inefficient, NS loss is commonly used for training KGE models. The NS loss in KGE, $\ell_{kge}(\mathbf{\theta})$ is represented as follows:

\begin{align}
    &\ell_{kge}(\mathbf{\theta})\nonumber \\
=&-\frac{1}{|D|}\sum_{(x,y) \in D} \Bigl[\log(\sigma(s_{\mathbf{\theta}}(x,y)+\gamma))\nonumber\\
    &+\frac{1}{\nu}\sum_{y_{i}\sim p_n(y_{i}|x)}^{\nu}\log(\sigma(-s_{\mathbf{\theta}}(x,y_i)-\gamma))\Bigr],
\label{eq:ns:kge}
\end{align}
where $\sigma$ is a sigmoid function, $p_n(y_{i}|x)$ is a noise distribution describing negative samples, $\nu$ is a number of negative samples per positive sample $(x, y)$, $\gamma$ is a margin term to adjust the value range of the score function. $p_n(y_i|x)$ has a role of adjusting the frequency of $y_{i}$ \cite{kamigaito-hayashi-2021-unified}. 

\subsection{Negative Sampling with Subsampling}
\label{subsec:ns_w_subsampling}

Subsampling \cite{ns} is a method to reduce the bias of training data by discounting high-frequent instances. 
\citet{pmlr-v162-kamigaito22a} show a general formulation to cover currently proposed subsampling approaches in the NS loss for KGE by altering two terms $A_{cbs}$ and $B_{cbs}$. In that form, the NS loss in KGE with subsampling, $\ell_{cbs}(\mathbf{\theta})$ is represented as follows:

\begin{align}
    &\ell_{cbs}(\mathbf{\theta}) \nonumber \\
=&-\frac{1}{|D|}\sum_{(x,y) \in D} \Bigl[A_{cbs}\log(\sigma(s_{\mathbf{\theta}}(x,y)+\gamma))\nonumber\\
    &+\frac{1}{\nu}\sum_{y_{i}\sim p_n(y_{i}|x)}^{\nu}B_{cbs}\log(\sigma(-s_{\mathbf{\theta}}(x,y_i)-\gamma))\Bigr],
\label{eq:subsamp}
\end{align}
where $A_{cbs}$ adjusts the frequency of a true link $(x, y)$, and $B_{cbs}$ adjusts the query $x$ to adjust the frequency of a false link $(x,y_i)$.

\begin{table}[t]
    \centering
    \resizebox{\columnwidth}{!}{
    \begin{tabular}{ccc}
    \toprule
         Method & $A_{cbs}$ & $B_{cbs}$  \\
         \midrule
         Base & $\frac{\frac{1}{\sqrt{\#(x,y)}}|D|}{\sum_{(x',y') \in D}\frac{1}{\sqrt{\#(x',y')}}}$ & $\frac{\frac{1}{\sqrt{\#(x,y)}}|D|}{\sum_{(x',y') \in D}\frac{1}{\sqrt{\#(x',y')}}}$ \\
         \midrule
         Freq & $\frac{\frac{1}{\sqrt{\#(x,y)}}|D|}{\sum_{(x',y') \in D}\frac{1}{\sqrt{\#(x',y')}}}$ & $\frac{\frac{1}{\sqrt{\#x}}|D|}{\sum_{x' \in D}{\frac{1}{\sqrt{\#x'}}}}$ \\
         \midrule
         Uniq & $\frac{\frac{1}{\sqrt{\#x}}|D|}{\sum_{x' \in D}{\frac{1}{\sqrt{\#x'}}}}$ & $\frac{\frac{1}{\sqrt{\#x}}|D|}{\sum_{x' \in D}{\frac{1}{\sqrt{\#x'}}}}$ \\
    \bottomrule 
    \end{tabular}
    }
    \caption{Currently proposed count-based subsampling methods in KGE and their corresponding terms on $A_{cbs}$ and $B_{cbs}$.}
    \label{tab:subsampling}
\end{table}

Table~\ref{tab:subsampling} lists the currently proposed subsampling approaches which are the original subsampling for word2vec \citep{ns} in KGE of \citet{rotate} (Base), frequency-based subsampling of \citet{pmlr-v162-kamigaito22a} (Freq), and unique-based subsampling of \citet{pmlr-v162-kamigaito22a} (Uniq) \cite{kamigaito2022subsampling}. Here, $\#$ denotes frequency, $\#(x,y)$ represents the frequency of $(x,y)$.

Since frequency for each link $(x,y)$ is at most one in KG, the previous approaches use the following back-off approximation \citep{katz1987estimation}:
\begin{align}
 \#(x,y) \approx \frac{\#(h_{i},r_{j})\!+\#(r_{j},t_{k})}{2},
 \label{eq:subsamp:approx}
\end{align}
where $(x,y)$ corresponds to the link $(h_{i}, r_{j}, t_{k})$, and $(h_{i},r_{j})$ and $(r_{j},t_{k})$ are the queries. Due to their heavily relying on counted frequency information of queries, we call the above conventional subsampling method \textbf{Count-based Subsamping} (\textbf{CBS}), hereafter. 

\section{Proposed Methods}

As shown in Equation~(\ref{eq:subsamp:approx}), CBS approximates the frequency of a link $\#(x,y)$ by combining the counted frequencies of entity-relation pairs. Thus, CBS cannot estimate $\#(x,y)$ well when at least one pair's frequency is low in the approximation. This kind of situation is caused by the sparseness problem in the KG datasets. 
To deal with this sparseness problem, we propose \textbf{Model-based Subsampling} method (\textbf{MBS}) and \textbf{Mixed Subsampling} method (\textbf{MIX}) as described in the following subsections.

\subsection{Model-based Subsampling (MBS)}
\label{subsec:mbs}

To avoid the problem caused by low-frequent entity-relation pairs, our MBS uses the estimated probabilities from a trained model $\mathbf{\theta}'$ to calculate frequencies for each triplet and query. By using $\mathbf{\theta}'$, the NS loss in KGE with MBS is represented as follows:
\begin{align}
    &\ell_{mbs}(\mathbf{\theta};\mathbf{\theta}') \nonumber \\
=&-\frac{1}{|D|}\sum_{(x,y) \in D} \Bigl[A_{mbs}(\mathbf{\theta}')\log(\sigma(s_{\mathbf{\theta}}(x,y)+\gamma))\nonumber\\
    &+\frac{1}{\nu}\!\!\!\!\!\!\sum_{y_{i}\sim p_n(y_{i}|x)}^{\nu}\!\!\!\!\!\!B_{mbs}(\mathbf{\theta}')\log(\sigma(-s_{\mathbf{\theta}}(x,y_i)-\gamma))\Bigr],
\label{eq:loss:mbs}
\end{align}
Here, corresponding to each method in Table \ref{tab:wn18rr}, $A_{mbs}(\mathbf{\theta}')$ and $B_{mbs}(\mathbf{\theta}')$ are further represented as follows:
\begin{align}
    A_{mbs}(\mathbf{\theta}') & = \left\{
\begin{array}{*2{>{\displaystyle}l}}
\frac{\#(x,y)_{mbs}^{-\alpha}|D|}{\sum_{(x',y') \in D}\#(x',y')_{mbs}^{-\alpha}} & \mathrm{(Base)} \\
\frac{\#(x,y)_{mbs}^{-\alpha}|D|}{\sum_{(x',y') \in D}\#(x',y')_{mbs}^{-\alpha}} & \mathrm{(Freq)} \\
\frac{\#x_{mbs}^{-\alpha}|D|}{\sum_{x'_{mbs} \in D}{{\#x'}_{mbs}^{-\alpha}}} & \mathrm{(Uniq)}
\label{eq:loss:mbs-a}
\end{array}
\right.
\end{align}
\begin{align}
    B_{mbs}(\mathbf{\theta}') & = \left\{
\begin{array}{*2{>{\displaystyle}l}}
\frac{\#(x,y)_{mbs}^{-\alpha}|D|}{\sum_{(x',y') \in D}\#(x',y')_{mbs}^{-\alpha}} & \mathrm{(Base)} \\
\frac{\#x_{mbs}^{-\alpha}|D|}{\sum_{x'_{mbs} \in D}{\#x'}_{mbs}^{-\alpha}} & \mathrm{(Freq)} \\
\frac{\#x_{mbs}^{-\alpha}|D|}{\sum_{x'_{mbs} \in D}{\#x'}_{mbs}^{-\alpha}} & \mathrm{(Uniq)}
\label{eq:loss:mbs-b}
\end{array}
\right.
\end{align}
where $\alpha$ is a temperature term to adjust the distribution on $A_{mbs}(\mathbf{\theta}')$ and $B_{mbs}(\mathbf{\theta}')$. The frequencies $\#(x,y)_{mbs}$ and $\#x_{mbs}$, estimated by using $score_{\mathbf{\theta}'}(x,y)$ are calculated as follows:
\begin{align}
    \#(x,y)_{mbs}& = |D|p_{\mathbf{\theta}'}(x,y),\\
    \#x_{mbs}& = |D|\textstyle\sum_{y_{i}\in D}p_{\mathbf{\theta}'}(x,y_{i}),\\
    p_{\mathbf{\theta}'}(x,y) &= \frac{e^{score_{\mathbf{\theta}'}(x,y)}}{\sum_{(x',y')\in D}e^{score_{\mathbf{\theta}'}(x',y')}}\label{eq:norm:score}.
\end{align}
Hereafter, we refer to a model pre-trained for MBS as a sub-model.
Different from the counted frequencies in Eq.~(\ref{eq:subsamp:approx}), $score_{\mathbf{\theta}'}(x,y)$ in Eq.~(\ref{eq:norm:score}) estimates them by sub-model inference regardless of their actual frequencies. Hence, we can expect MBS to deal with the sparseness problem in CBS. However, the ability of MBS depends on the sub-model, and we investigated the performance through our evaluations (\S\ref{sec:evaluation}). 

\subsection{Mixed Subsampling (MIX)}
\label{subsec:mix}

As discussed in language modeling context \citep{neubig-dyer-2016-generalizing}, count-based and model-based frequencies have different strengths and weaknesses. To boost the advantages of CBS and MBS by mitigating their disadvantages, MIX uses a mixture of the distribution as follows:
\begin{align}
    &\ell_{mix}(\mathbf{\theta};\mathbf{\theta}') \nonumber \\
=&-\frac{1}{|D|}\sum_{(x,y) \in D} \Bigl[A_{mix}(\mathbf{\theta}')\log(\sigma(s_{\mathbf{\theta}}(x,y)+\gamma))\nonumber\\
    &+\frac{1}{\nu}\!\!\!\!\!\!\sum_{y_{i}\sim p_n(y_{i}|x)}^{\nu}\!\!\!\!\!\!B_{mix}(\mathbf{\theta}')\log(\sigma(-s_{\mathbf{\theta}}(x,y_i)-\gamma))\Bigr],
\label{eq:loss:mix}
\end{align}
where $A_{mix}(\mathbf{\theta}')$ is a mixture of $A_{cbs}$ in Eq.~(\ref{eq:subsamp}) and $A_{mbs}(\mathbf{\theta}')$ in Eq.~(\ref{eq:loss:mbs}), and $B_{mix}(\mathbf{\theta}')$ is also a mixture of  $B_{cbs}$ in Eq.~(\ref{eq:subsamp}) and $B_{mbs}(\mathbf{\theta}')$ Eq.~(\ref{eq:loss:mbs}) as follows: 
\begin{align}
    A_{mix}(\mathbf{\theta}') = \lambda A_{mbs}(\mathbf{\theta}') + (1-\lambda)A_{cbs}  \label{eq:loss:mix-a}\\
    B_{mix}(\mathbf{\theta}') = \lambda B_{mbs}(\mathbf{\theta}') + (1-\lambda)B_{cbs} 
    \label{eq:loss:mix-b}
\end{align}
where $\lambda$ is a hyper-parameter to adjust the ratio of MBS and CBS.
Note that MIX can be interpreted as a kind of multi-task learning\footnote{See Appendix \ref{appendix:mix-multi-task} for the details.}.

\section{Evaluation and Analysis}
\label{sec:evaluation}

\subsection{Settings}
\label{sec:setting}

\paragraph{Datasets} 
We used the three commonly used datasets, FB15k-237, WN18RR, and YAGO3-10, for the evaluation. Table \ref{tab:datasets} shows the statistics for each dataset. Unlike FB15k-237 and WN18RR, the dataset of YAGO3-10 only includes entities that have at least 10 relations and alleviates the sparseness problem of KGs. Thus, we can investigate the effectiveness of MBS and MIX in the sparseness problem by comparing performances on these datasets.

\paragraph{Methods} 

We compared five popular KGE models RotatE, TransE, HAKE, ComplEx, and DistMult with utilizing subsampling methods Base, Freq, and Uniq based on the loss of CBS (\S\ref{subsec:ns_w_subsampling}) and our MBS (\S\ref{subsec:mbs}) and MIX (\S\ref{subsec:mix}).
Additionally, we conducted experiments with no subsampling (None) to investigate the efficacy of the subsampling method. In YAGO3-10, due to our limited computational resources and the existence of tuned hyper-parameters by \citet{rotate,https://doi.org/10.48550/arxiv.1911.09419}, we only used RotatE and HAKE for evaluation.

\begin{table}[t]
\centering								
\resizebox{1\columnwidth}{!}{								
    \begin{tabular}{lrrrrrr}								
	\toprule
        \textbf{Dataset}	&\textbf{\#Train}	&\textbf{\#Valid}	&\textbf{\#Test}	&\textbf{Ent}	&\textbf{Rel}\\
        \midrule
	FB15K-237	&272,115	&17,535     &20,466     &14,541     &237 \\
	WN18RR	    &86,835	    &3,034	    &3,134	    &40,943	    &11	\\
	YAGO3-10	&1,079,040	&5,000	    &5,000	    &123,188	&37	 \\
	\bottomrule	
    \end{tabular}}					
\caption{Datasets statistics. \#: Split in terms of number of triples; Ent: Entities; Rel: Relations; Exa: Examples.}
\label{tab:datasets}
\end{table}								
\begin{table*}[htbp]
    \centering
    \resizebox{0.9\textwidth}{!}{
    \small
    {
    \renewcommand{\arraystretch}{0.9}
    \begin{tabular}{ccccccccccccccc}
    \toprule
    \multicolumn{15}{c}{\textbf{FB15k-237}} \\
    \midrule
    \multirow{2}[1]{*}{\textbf{Model}} & \multicolumn{2}{c}{\multirow{2}[1]{*}{\textbf{Subsampling}}} & \multicolumn{2}{c}{\textbf{MRR}} & \multicolumn{2}{c}{\textbf{H@1}} & \multicolumn{2}{c}{\textbf{H@3}} & \multicolumn{2}{c}{\textbf{H@10}} & \multicolumn{4}{c}{\textbf{Submodeling}} \\
\cmidrule{4-15}          & \multicolumn{2}{c}{} & \textbf{Mean} & \textbf{SD} & \textbf{Mean} & \textbf{SD} & \textbf{Mean} & \textbf{SD} & \textbf{Mean} & \textbf{SD} & \multicolumn{2}{c}{\textbf{Sub-model}} & \textbf{$\alpha$} & \textbf{$\lambda$} \\
    \midrule
    \multirow{7.5}[20]{*}{RotatE} & \multicolumn{2}{c}{None} & 32.9  & 0.1   & 22.9  & 0.1   & 37.0  & 0.1   & 53.1  & 0.1   &       &       &       &  \\
\cmidrule{2-15}          & \multirow{1.5}[6]{*}{Base} & CBS   & 33.6  & 0.1   & 23.9  & 0.1   & 37.4  & 0.1   & 53.1  & 0.1   &       &       &       &  \\
\cmidrule{3-15}          &       & MBS   & \textbf{33.9} & 0.0   & \textbf{24.2} & 0.1   & \textbf{37.7} & 0.1   & \textbf{53.5} & 0.1   & \multirow{2}[1]{*}{ComplEx} & \multirow{2}[1]{*}{None} & \multirow{2}[1]{*}{0.5} & -- \\
                         &       & MIX   & \textbf{33.9} & 0.0   & \textbf{24.2} & 0.0   & \textbf{37.7} & 0.1   & \textbf{53.5} & 0.1   &       &       &       & 0.9 \\
\cmidrule{2-15}          & \multirow{1.5}[6]{*}{Freq} & CBS   & 34.1  & 0.0   & 24.6  & 0.1   & 37.7  & 0.0   & 53.1  & 0.0   &       &       &       &  \\
\cmidrule{3-15}          &       & MBS   & 34.3  & 0.0   & 24.8  & 0.1   & 38.0  & 0.1   & 53.6  & 0.1   & \multirow{2}[1]{*}{ComplEx} & \multirow{2}[1]{*}{None} & \multirow{2}[1]{*}{0.1} & -- \\
                         &       & MIX   & $^{\dagger}${\textbf{34.5}} & 0.0   & \textbf{24.9} & 0.1   & $^{\dagger}${\textbf{38.1}} & 0.1   & $^{\dagger}${\textbf{53.7}} & 0.1   &       &       &       & 0.7 \\
\cmidrule{2-15}          & \multirow{1.5}[6]{*}{Uniq} & CBS   & 33.9  & 0.0   & 24.4  & 0.1   & 37.6  & 0.1   & 53.0  & 0.2   &       &       &       &  \\
\cmidrule{3-15}          &       & MBS   & 34.3  & 0.0   & 24.7  & 0.1   & \textbf{38.0} & 0.2   & $^{\dagger}${\textbf{53.7}} & 0.1   & \multirow{2}[1]{*}{ComplEx} & \multirow{2}[1]{*}{None} & \multirow{2}[1]{*}{0.1} & -- \\
                         &       & MIX   & $^{\dagger}${\textbf{34.5}} & 0.1   & $^{\dagger}${\textbf{25.0}} & 0.1   & \textbf{38.0} & 0.1   & 53.6  & 0.1   &       &       &       & 0.5 \\
    \midrule
    \multirow{7.5}[20]{*}{TransE} & \multicolumn{2}{c}{None} & 33.0  & 0.1   & 22.9  & 0.1   & 37.2  & 0.1   & 53.0  & 0.2   &       &       &       &  \\
\cmidrule{2-15}          & \multirow{1.5}[6]{*}{Base} & CBS   & 33.0  & 0.1   & 23.1  & 0.1   & 36.8  & 0.1   & 52.7  & 0.1   &       &       &       &  \\
\cmidrule{3-15}          &       & MBS   & 33.3  & 0.1   & \textbf{23.4} & 0.1   & \textbf{37.3} & 0.0   & 53.1  & 0.1   & \multirow{2}[1]{*}{ComplEx} & \multirow{2}[1]{*}{None} & \multirow{2}[1]{*}{0.5} & -- \\
                         &       & MIX   & \textbf{33.4} & 0.1   & \textbf{23.4} & 0.1   & \textbf{37.3} & 0.0   & \textbf{53.2} & 0.1   &       &       &       & 0.9 \\
\cmidrule{2-15}          & \multirow{1.5}[6]{*}{Freq} & CBS   & 33.5  & 0.1   & 23.9  & 0.2   & 37.3  & 0.1   & 52.8  & 0.1   &       &       &       &  \\
\cmidrule{3-15}          &       & MBS   & 33.9  & 0.0   & 24.1  & 0.1   & 37.7  & 0.1   & 53.2  & 0.0   & \multirow{2}[1]{*}{RotatE} & \multirow{2}[1]{*}{Base} & \multirow{2}[1]{*}{0.1} & -- \\
                         &       & MIX   & $^{\dagger}${\textbf{34.1}} & 0.1   & $^{\dagger}${\textbf{24.3}} & 0.1   & $^{\dagger}${\textbf{37.9}} & 0.1   & $^{\dagger}${\textbf{53.4}} & 0.0   &       &       &       & 0.7 \\
\cmidrule{2-15}          & \multirow{1.5}[6]{*}{Uniq} & CBS   & 33.6  & 0.1   & 24.0  & 0.1   & 37.3  & 0.1   & 52.7  & 0.1   &       &       &       &  \\
\cmidrule{3-15}          &       & MBS   & 33.8  & 0.1   & 23.9  & 0.1   & 37.7  & 0.1   & 53.2  & 0.1   & \multirow{2}[1]{*}{RotatE} & \multirow{2}[1]{*}{Base} & \multirow{2}[1]{*}{0.1} & -- \\
                         &       & MIX   & \textbf{34.0} & 0.1   & $^{\dagger}${\textbf{24.3}} & 0.1   & $^{\dagger}${\textbf{37.9}} & 0.1   & $^{\dagger}${\textbf{53.4}} & 0.1   &       &       &       & 0.7 \\
    \midrule
    \multirow{7.5}[20]{*}{HAKE} & \multicolumn{2}{c}{None} & 32.4  & 0.0   & 22.0  & 0.0   & 36.9  & 0.1   & 53.1  & 0.1   &       &       &       &  \\
\cmidrule{2-15}          & \multirow{1.5}[6]{*}{Base} & CBS   & 34.5  & 0.1   & \textbf{24.9} & 0.1   & \textbf{38.3} & 0.1   & 54.1  & 0.0   &       &       &       &  \\
\cmidrule{3-15}          &       & MBS   & 34.4  & 0.1   & 24.5  & 0.1   & 38.2  & 0.1   & 54.1  & 0.2   & \multirow{2}[1]{*}{ComplEx} & \multirow{2}[1]{*}{None} & \multirow{2}[1]{*}{1.0} & -- \\
                         &       & MIX   & \textbf{34.6} & 0.1   & 24.8  & 0.1   & \textbf{38.3} & 0.1   & \textbf{54.2} & 0.1   &       &       &       & 0.5 \\
\cmidrule{2-15}          & \multirow{1.5}[6]{*}{Freq} & CBS   & 35.1  & 0.1   & 25.5  & 0.1   & 38.8  & 0.1   & 54.3  & 0.0   &       &       &       &  \\
\cmidrule{3-15}          &       & MBS   & 35.2  & 0.1   & 25.7  & 0.1   & 39.0  & 0.1   & 54.4  & 0.1   & \multirow{2}[1]{*}{ComplEx} & \multirow{2}[1]{*}{Base} & \multirow{2}[1]{*}{0.5} &  -- \\
                         &       & MIX   & $^{\dagger}${\textbf{35.4}} & 0.0   & $^{\dagger}${\textbf{25.8}} & 0.1   & $^{\dagger}${\textbf{39.2}} & 0.1   & \textbf{54.5} & 0.1   &       &       &       & 0.5 \\
\cmidrule{2-15}          & \multirow{1.5}[6]{*}{Uniq} & CBS   & 35.2  & 0.1   & 25.6  & 0.2   & \textbf{38.9} & 0.1   & 54.5  & 0.1   &       &       &       &  \\
\cmidrule{3-15}          &       & MBS   & \textbf{35.3} & 0.1   & 25.6  & 0.0   & \textbf{38.9} & 0.1   & 54.5  & 0.1   & \multirow{2}[1]{*}{RotatE} & \multirow{2}[1]{*}{Base} & \multirow{2}[1]{*}{0.5} & --  \\
                         &       & MIX   & \textbf{35.3} & 0.0   & $^{\dagger}${\textbf{25.8}} & 0.1   & \textbf{38.9} & 0.1   & $^{\dagger}${\textbf{54.6}} & 0.1   &       &       &       & 0.3 \\
    \midrule
    \multirow{7.5}[20]{*}{ComplEx} & \multicolumn{2}{c}{None} & 22.3  & 0.1   & 13.9  & 0.1   & 24.1  & 0.2   & 39.5  & 0.1   &       &       &       &  \\
\cmidrule{2-15}          & \multirow{1.5}[6]{*}{Base} & CBS   & 32.3  & 0.1   & \textbf{23.0} & 0.2   & 35.5  & 0.1   & 51.3  & 0.1   &       &       &       &  \\
\cmidrule{3-15}          &       & MBS   & 31.2  & 0.1   & 21.7  & 0.1   & 34.4  & 0.2   & 50.6  & 0.1   & \multirow{2}[1]{*}{ComplEx} & \multirow{2}[1]{*}{None} & \multirow{2}[1]{*}{1.0} &  -- \\
                         &       & MIX   & \textbf{32.4} & 0.1   & 22.8  & 0.1   & \textbf{35.8} & 0.1   & $^{\dagger}${\textbf{52.1}} & 0.2   &       &       &       & 0.5 \\
\cmidrule{2-15}          & \multirow{1.5}[6]{*}{Freq} & CBS   & 32.7  & 0.1   & 23.6  & 0.1   & 36.0  & 0.1   & 51.2  & 0.1   &       &       &       &  \\
\cmidrule{3-15}          &       & MBS   & 32.0  & 0.0   & 23.0  & 0.0   & 35.1  & 0.1   & 50.1  & 0.1   & \multirow{2}[1]{*}{DistMult} & \multirow{2}[1]{*}{Base} & \multirow{2}[1]{*}{0.5} & -- \\
                         &       & MIX   & $^{\dagger}${\textbf{32.9}} & 0.1   & $^{\dagger}${\textbf{23.7}} & 0.1   & $^{\dagger}${\textbf{36.2}} & 0.1   & \textbf{51.3} & 0.2   &       &       &       & 0.1 \\
\cmidrule{2-15}          & \multirow{1.5}[6]{*}{Uniq} & CBS   & 32.6  & 0.1   & \textbf{23.4} & 0.2   & 35.9  & 0.1   & 51.1  & 0.1   &       &       &       &  \\
\cmidrule{3-15}          &       & MBS   & 31.8  & 0.1   & 22.6  & 0.2   & 34.9  & 0.1   & 50.5  & 0.2   & \multirow{2}[1]{*}{ComplEx} & \multirow{2}[1]{*}{Base} & \multirow{2}[1]{*}{0.5} &  -- \\
                         &       & MIX   & \textbf{32.7} & 0.1   & \textbf{23.4} & 0.1   & \textbf{36.0} & 0.1   & \textbf{51.2} & 0.2   &       &       &       & 0.1 \\
    \midrule
    \multirow{7.5}[20]{*}{DistMult} & \multicolumn{2}{c}{None} & 22.3  & 0.1   & 14.1  & 0.1   & 24.2  & 0.1   & 39.3  & 0.1   &       &       &       &  \\
\cmidrule{2-15}          & \multirow{1.5}[6]{*}{Base} & CBS   & 30.8  & 0.1   & 22.0  & 0.1   & 33.7  & 0.1   & 48.4  & 0.1   &       &       &       &  \\
\cmidrule{3-15}          &       & MBS   & 31.1  & 0.2   & 21.8  & 0.1   & 34.1  & 0.2   & 49.6  & 0.2   & \multirow{2}[1]{*}{ComplEx} & \multirow{2}[1]{*}{None} & \multirow{2}[1]{*}{1.0} &  -- \\
                         &       & MIX   & $^{\dagger}${\textbf{31.3}} & 0.1   & $^{\dagger}${\textbf{22.3}} & 0.1   & $^{\dagger}${\textbf{34.3}} & 0.1   & $^{\dagger}${\textbf{49.7}} & 0.1   &       &       &       & 0.7 \\
\cmidrule{2-15}          & \multirow{1.5}[6]{*}{Freq} & CBS   & \textbf{29.9} & 0.1   & \textbf{21.2} & 0.1   & \textbf{32.8} & 0.1   & \textbf{47.5} & 0.0   &       &       &       &  \\
\cmidrule{3-15}          &       & MBS   & 27.9  & 0.1   & 19.6  & 0.2   & 30.4  & 0.2   & 44.4  & 0.1   & \multirow{2}[1]{*}{DistMult} & \multirow{2}[1]{*}{Base} & \multirow{2}[1]{*}{0.5} &  -- \\
                         &       & MIX   & 29.7  & 0.1   & 20.9  & 0.1   & 32.6  & 0.1   & \textbf{47.5} & 0.1   &       &       &       & 0.1 \\
\cmidrule{2-15}          & \multirow{1.5}[6]{*}{Uniq} & CBS   & \textbf{29.2} & 0.0   & \textbf{20.4} & 0.1   & \textbf{31.9} & 0.0   & \textbf{46.7} & 0.1   &       &       &       &  \\
\cmidrule{3-15}          &       & MBS   & 27.9  & 0.1   & 19.3  & 0.0   & 30.3  & 0.1   & 45.2  & 0.1   & \multirow{2}[1]{*}{ComplEx} & \multirow{2}[1]{*}{Base} & \multirow{2}[1]{*}{0.5} &  -- \\
                         &       & MIX   & 29.1  & 0.0   & 20.3  & 0.1   & 31.8  & 0.1   & 46.6  & 0.1   &       &       &       & 0.1 \\
    \bottomrule
    \end{tabular}}
    }%
    \caption{Results on FB15k-237. The bold scores are the best results for each subsampling type (e.g. \textit{Base}, \textit{Freq}, and \textit{Uniq}.). $\dagger$ indicates the best scores for each model. \textit{SD} denotes the standard deviation of the three trial. \textit{Sub-model}, $\alpha$, and $\lambda$ denote the sub-model, temperature, and mixing ratio chosen by development data. \label{tab:fb15k-237}}
\end{table*}%

\begin{table*}[htbp]
    \centering
    \resizebox{0.9\textwidth}{!}{
    \small
    {
    \renewcommand{\arraystretch}{0.9}
    \begin{tabular}{ccccccccccccccc}
    \toprule
    \multicolumn{15}{c}{\textbf{WN18RR}} \\
    \midrule
    \multirow{2}[1]{*}{\textbf{Model}} & \multicolumn{2}{c}{\multirow{2}[1]{*}{\textbf{Subsampling}}} & \multicolumn{2}{c}{\textbf{MRR}} & \multicolumn{2}{c}{\textbf{H@1}} & \multicolumn{2}{c}{\textbf{H@3}} & \multicolumn{2}{c}{\textbf{H@10}} & \multicolumn{4}{c}{\textbf{Submodeling}} \\
\cmidrule{4-15}          & \multicolumn{2}{c}{} & \textbf{Mean} & \textbf{SD} & \textbf{Mean} & \textbf{SD} & \textbf{Mean} & \textbf{SD} & \textbf{Mean} & \textbf{SD} & \multicolumn{2}{c}{\textbf{Sub-model}} & \textbf{$\alpha$} & \textbf{$\lambda$} \\
    \midrule
    \multirow{7.5}[20]{*}{RotatE} & \multicolumn{2}{c}{None} & 47.3  & 0.1   & 42.9  & 0.4   & 48.8  & 0.3   & 55.7  & 0.7   &       &       &       &  \\
\cmidrule{2-15}          & \multirow{1.5}[6]{*}{Base} & CBS   & 47.6  & 0.1   & $^{\dagger}${\textbf{43.3}} & 0.2   & 49.3  & 0.3   & 56.1  & 0.5   &       &       &       &  \\
\cmidrule{3-15}          &       & MBS   & $^{\dagger}${\textbf{48.0}} & 0.0   & $^{\dagger}${\textbf{43.3}} & 0.2   & \textbf{49.6} & 0.2   & $^{\dagger}${\textbf{57.5}} & 0.4   & \multirow{2}[1]{*}{ComplEx} & \multirow{2}[1]{*}{None} & \multirow{2}[1]{*}{1.0} & -- \\
                         &       & MIX   & 47.8  & 0.1   & 43.2  & 0.2   & 49.5  & 0.2   & 57.2  & 0.3   &       &       &       & 0.5 \\
\cmidrule{2-15}          & \multirow{1.5}[6]{*}{Freq} & CBS   & 47.7  & 0.1   & \textbf{43.2} & 0.3   & 49.5  & 0.3   & 56.9  & 0.9   &       &       &       &  \\
\cmidrule{3-15}          &       & MBS   & \textbf{47.9} & 0.1   & \textbf{43.2} & 0.2   & 49.6  & 0.2   & 57.4  & 0.4   & \multirow{2}[1]{*}{ComplEx} & \multirow{2}[1]{*}{None} & \multirow{2}[1]{*}{0.5} & -- \\
                         &       & MIX   & \textbf{47.9} & 0.1   & 42.9  & 0.1   & \textbf{49.8} & 0.1   & $^{\dagger}${\textbf{57.5}} & 0.2   &       &       &       & 0.3 \\
\cmidrule{2-15}          & \multirow{1.5}[6]{*}{Uniq} & CBS   & 47.7  & 0.1   & 43.1  & 0.1   & 49.6  & 0.2   & 56.9  & 0.4   &       &       &       &  \\
\cmidrule{3-15}          &       & MBS   & $^{\dagger}${\textbf{48.0}} & 0.1   & \textbf{43.2} & 0.2   & $^{\dagger}${\textbf{49.9}} & 0.2   & $^{\dagger}${\textbf{57.5}} & 0.2   & \multirow{2}[1]{*}{ComplEx} & \multirow{2}[1]{*}{None} & \multirow{2}[1]{*}{0.5} & -- \\
                         &       & MIX   & 47.8  & 0.1   & 43.0  & 0.1   & 49.7  & 0.3   & 57.2  & 0.5   &       &       &       & 0.5 \\
    \midrule
    \multirow{7.5}[20]{*}{TransE} & \multicolumn{2}{c}{None} & 22.5  & 0.0   & 1.7   & 0.0   & 40.1  & 0.1   & 52.5  & 0.2   &       &       &       &  \\
\cmidrule{2-15}          & \multirow{1.5}[6]{*}{Base} & CBS   & 22.3  & 0.1   & 1.3   & 0.1   & 40.1  & 0.2   & 53.0  & 0.0   &       &       &       &  \\
\cmidrule{3-15}          &       & MBS   & \textbf{23.7} & 0.1   & \textbf{2.5} & 0.1   & 41.2  & 0.2   & 53.1  & 0.1   & \multirow{2}[1]{*}{ComplEx} & \multirow{2}[1]{*}{Base} & \multirow{2}[1]{*}{2.0} & -- \\
                         &       & MIX   & 23.6  & 0.1   & 2.4   & 0.1   & \textbf{41.4} & 0.1   & \textbf{53.2} & 0.2   &       &       &       & 0.9 \\
\cmidrule{2-15}          & \multirow{1.5}[6]{*}{Freq} & CBS   & 23.0  & 0.0   & 1.9   & 0.1   & 40.9  & 0.1   & 53.7  & 0.0   &       &       &       &  \\
\cmidrule{3-15}          &       & MBS   & $^{\dagger}${\textbf{25.0}} & 0.1   & $^{\dagger}${\textbf{4.2}} & 0.1   & 42.4  & 0.2   & 54.1  & 0.0   & \multirow{2}[1]{*}{ComplEx} & \multirow{2}[1]{*}{Base} & \multirow{2}[1]{*}{2.0} & -- \\
                         &       & MIX   & $^{\dagger}${\textbf{25.0}} & 0.1   & 4.0   & 0.2   & $^{\dagger}${\textbf{42.6}} & 0.1   & $^{\dagger}${\textbf{54.3}} & 0.1   &       &       &       & 0.9 \\
\cmidrule{2-15}          & \multirow{1.5}[6]{*}{Uniq} & CBS   & 23.2  & 0.1   & 2.2   & 0.1   & 40.9  & 0.2   & 53.6  & 0.2   &       &       &       &  \\
\cmidrule{3-15}          &       & MBS   & \textbf{23.9} & 0.1   & \textbf{3.3} & 0.1   & 40.8  & 0.1   & \textbf{54.2} & 0.1   & \multirow{2}[1]{*}{ComplEx} & \multirow{2}[1]{*}{Base} & \multirow{2}[1]{*}{1.0} & -- \\
                         &       & MIX   & \textbf{23.9} & 0.1   & \textbf{3.3} & 0.0   & \textbf{41.1} & 0.2   & \textbf{54.2} & 0.2   &       &       &       & 0.9 \\
    \midrule
    \multirow{7.5}[20]{*}{HAKE} & \multicolumn{2}{c}{None} & 49.0  & 0.1   & 44.6  & 0.2   & 50.7  & 0.1   & 57.5  & 0.2   &       &       &       &  \\
\cmidrule{2-15}          & \multirow{1.5}[6]{*}{Base} & CBS   & \textbf{49.6} & 0.0   & \textbf{45.1} & 0.2   & \textbf{51.5} & 0.2   & \textbf{58.2} & 0.1   &       &       &       &  \\
\cmidrule{3-15}          &       & MBS   & 49.2  & 0.1   & 44.7  & 0.1   & 51.0  & 0.3   & 58.0  & 0.1   & \multirow{2}[1]{*}{ComplEx} & \multirow{2}[1]{*}{None} & \multirow{2}[1]{*}{0.5} & -- \\
                         &       & MIX   & 49.5  & 0.1   & 45.0  & 0.2   & 51.4  & 0.2   & \textbf{58.2} & 0.1   &       &       &       & 0.1 \\
\cmidrule{2-15}          & \multirow{1.5}[6]{*}{Freq} & CBS   & 49.7  & 0.0   & 45.1  & 0.1   & 51.5  & 0.2   & 58.4  & 0.2   &       &       &       &  \\
\cmidrule{3-15}          &       & MBS   & $^{\dagger}${\textbf{49.9}} & 0.1   & $^{\dagger}${\textbf{45.4}} & 0.1   & \textbf{51.7} & 0.2   & $^{\dagger}${\textbf{58.5}} & 0.1   & \multirow{2}[1]{*}{ComplEx} & \multirow{2}[1]{*}{None} & \multirow{2}[1]{*}{0.5} & -- \\
                         &       & MIX   & $^{\dagger}${\textbf{49.9}} & 0.1   & $^{\dagger}${\textbf{45.4}} & 0.1   & \textbf{51.7} & 0.2   & 58.4  & 0.3   &       &       &       & 0.9 \\
\cmidrule{2-15}          & \multirow{1.5}[6]{*}{Uniq} & CBS   & 49.7  & 0.1   & 45.2  & 0.2   & 51.6  & 0.2   & $^{\dagger}${\textbf{58.5}} & 0.3   &       &       &       &  \\
\cmidrule{3-15}          &       & MBS   & $^{\dagger}${\textbf{49.9}} & 0.1   & $^{\dagger}${\textbf{45.4}} & 0.1   & $^{\dagger}${\textbf{51.8}} & 0.2   & $^{\dagger}${\textbf{58.5}} & 0.1   & \multirow{2}[1]{*}{DistMult} & \multirow{2}[1]{*}{None} & \multirow{2}[1]{*}{0.5} & -- \\
                         &       & MIX   & $^{\dagger}${\textbf{49.9}} & 0.1   & $^{\dagger}${\textbf{45.4}} & 0.2   & $^{\dagger}${\textbf{51.8}} & 0.2   & $^{\dagger}${\textbf{58.5}} & 0.1   &       &       &       & 0.7 \\
    \midrule
    \multirow{7.5}[20]{*}{ComplEx} & \multicolumn{2}{c}{None} & 45.0  & 0.1   & 40.9  & 0.1   & 46.6  & 0.2   & 53.5  & 0.2   &       &       &       &  \\
\cmidrule{2-15}          & \multirow{1.5}[6]{*}{Base} & CBS   & 46.9  & 0.1   & 42.6  & 0.1   & 48.7  & 0.2   & 55.3  & 0.2   &       &       &       &  \\
\cmidrule{3-15}          &       & MBS   & \textbf{47.3} & 0.2   & \textbf{43.4} & 0.1   & \textbf{49.1} & 0.1   & \textbf{55.5} & 0.4   & \multirow{2}[1]{*}{ComplEx} & \multirow{2}[1]{*}{None} & \multirow{2}[1]{*}{2.0} & -- \\
                         &       & MIX   & \textbf{47.3} & 0.2   & \textbf{43.4} & 0.1   & \textbf{49.1} & 0.1   & \textbf{55.5} & 0.4   &       &       &       & 0.7 \\
\cmidrule{2-15}          & \multirow{1.5}[6]{*}{Freq} & CBS   & 47.3  & 0.2   & 43.0  & 0.2   & 49.2  & 0.2   & 56.1  & 0.2   &       &       &       &  \\
\cmidrule{3-15}          &       & MBS   & $^{\dagger}${\textbf{48.5}} & 0.1   & $^{\dagger}${\textbf{44.6}} & 0.1   & 49.9  & 0.3   & 56.5  & 0.2   & \multirow{2}[1]{*}{ComplEx} & \multirow{2}[1]{*}{None} & \multirow{2}[1]{*}{0.5} & -- \\
                         &       & MIX   & 48.4  & 0.2   & 44.4  & 0.1   & $^{\dagger}${\textbf{50.1}} & 0.1   & $^{\dagger}${\textbf{56.7}} & 0.4   &       &       &       & 0.9 \\
\cmidrule{2-15}          & \multirow{1.5}[6]{*}{Uniq} & CBS   & 47.5  & 0.2   & 43.1  & 0.2   & 49.4  & 0.1   & 56.1  & 0.2   &       &       &       &  \\
\cmidrule{3-15}          &       & MBS   & \textbf{48.4} & 0.1   & \textbf{44.3} & 0.1   & \textbf{50.0} & 0.2   & 56.5  & 0.1   & \multirow{2}[1]{*}{ComplEx} & \multirow{2}[1]{*}{None} & \multirow{2}[1]{*}{0.5} & -- \\
                         &       & MIX   & \textbf{48.4} & 0.1   & 44.2  & 0.2   & \textbf{50.0} & 0.2   & \textbf{56.6} & 0.2   &       &       &       & 0.9 \\
    \midrule
    \multirow{7.5}[20]{*}{DistMult} & \multicolumn{2}{c}{None} & 42.5  & 0.1   & 38.3  & 0.1   & 43.6  & 0.0   & 51.2  & 0.1   &       &       &       &  \\
\cmidrule{2-15}          & \multirow{1.5}[6]{*}{Base} & CBS   & 43.9  & 0.1   & 39.3  & 0.1   & 45.4  & 0.1   & 53.3  & 0.2   &       &       &       &  \\
\cmidrule{3-15}          &       & MBS   & 44.0  & 0.1   & 40.0  & 0.1   & 44.9  & 0.2   & 52.4  & 0.4   & \multirow{2}[1]{*}{ComplEx} & \multirow{2}[1]{*}{None} & \multirow{2}[1]{*}{2.0} & -- \\
                         &       & MIX   & \textbf{44.6} & 0.1   & \textbf{40.5} & 0.1   & \textbf{45.7} & 0.3   & \textbf{53.7} & 0.2   &       &       &       & 0.7 \\
\cmidrule{2-15}          & \multirow{1.5}[6]{*}{Freq} & CBS   & 44.5  & 0.1   & 39.9  & 0.2   & 46.0  & 0.2   & 54.3  & 0.2   &       &       &       &  \\
\cmidrule{3-15}          &       & MBS   & $^{\dagger}${\textbf{45.5}} & 0.1   & $^{\dagger}${\textbf{41.2}} & 0.2   & 46.6  & 0.2   & 54.6  & 0.1   & \multirow{2}[1]{*}{ComplEx} & \multirow{2}[1]{*}{None} & \multirow{2}[1]{*}{0.5} & -- \\
                         &       & MIX   & $^{\dagger}${\textbf{45.5}} & 0.1   & $^{\dagger}${\textbf{41.2}} & 0.1   & $^{\dagger}${\textbf{46.7}} & 0.3   & $^{\dagger}${\textbf{54.7}} & 0.1   &       &       &       & 0.9 \\
\cmidrule{2-15}          & \multirow{1.5}[6]{*}{Uniq} & CBS   & 44.8  & 0.1   & 40.1  & 0.2   & 46.3  & 0.3   & \textbf{54.5} & 0.2   &       &       &       &  \\
\cmidrule{3-15}          &       & MBS   & \textbf{45.3} & 0.1   & \textbf{41.1} & 0.2   & \textbf{46.4} & 0.1   & 54.3  & 0.1   & \multirow{2}[1]{*}{ComplEx} & \multirow{2}[1]{*}{None} & \multirow{2}[1]{*}{0.5} & -- \\
                         &       & MIX   & \textbf{45.3} & 0.1   & 41.0  & 0.2   & \textbf{46.4} & 0.1   & 54.4  & 0.2   &       &       &       & 0.9 \\
    \bottomrule
    \end{tabular}}
    }%
    \caption{Results on WN18RR. The notations are the same as the ones in Table \ref{tab:fb15k-237}.\label{tab:wn18rr}}
\end{table*}%

\begin{table*}[htbp]
    \centering
    \resizebox{0.9\textwidth}{!}{
    \small
    {
    \renewcommand{\arraystretch}{0.9}
    \begin{tabular}{ccccccccccccccc}
    \toprule
    \multicolumn{15}{c}{\textbf{YAGO3-10}} \\
    \midrule
    \multirow{2}[1]{*}{\textbf{Model}} & \multicolumn{2}{c}{\multirow{2}[1]{*}{\textbf{Subsampling}}} & \multicolumn{2}{c}{\textbf{MRR}} & \multicolumn{2}{c}{\textbf{H@1}} & \multicolumn{2}{c}{\textbf{H@3}} & \multicolumn{2}{c}{\textbf{H@10}} & \multicolumn{4}{c}{\textbf{Submodeling}} \\
\cmidrule{4-15}          & \multicolumn{2}{c}{} & \textbf{Mean} & \textbf{SD} & \textbf{Mean} & \textbf{SD} & \textbf{Mean} & \textbf{SD} & \textbf{Mean} & \textbf{SD} & \multicolumn{2}{c}{\textbf{Sub-model}} & \textbf{$\alpha$} & \textbf{$\lambda$} \\
    \midrule
    \multirow{7.5}[20]{*}{RotatE} & \multicolumn{2}{c}{None} & 49.2  & 0.2   & 39.6  & 0.2   & 55.0  & 0.2   & 67.2  & 0.3   &       &       &       &  \\
\cmidrule{2-15}          & \multirow{1.5}[6]{*}{Base} & CBS   & 49.3  & 0.1   & 39.9  & 0.1   & 54.9  & 0.3   & 67.1  & 0.2   &       &       &       &  \\
\cmidrule{3-15}          &       & MBS   & 49.5  & 0.2   & 40.0  & 0.3   & 55.4  & 0.0   & 66.8  & 0.2   & \multirow{2}[1]{*}{RotatE} & \multirow{2}[1]{*}{None} & \multirow{2}[1]{*}{0.5} & -- \\
                         &       & MIX   & \textbf{49.8} & 0.1   & \textbf{40.4} & 0.2   & \textbf{55.6} & 0.2   & \textbf{67.2} & 0.3   &       &       &       & 0.7 \\
\cmidrule{2-15}          & \multirow{1.5}[6]{*}{Freq} & CBS   & 49.6  & 0.1   & 40.2  & 0.1   & 55.2  & 0.1   & 67.3  & 0.1   &       &       &       &  \\
\cmidrule{3-15}          &       & MBS   & 50.1  & 0.2   & $^{\dagger}${\textbf{41.0}} & 0.2   & 55.6  & 0.2   & 67.1  & 0.1   & \multirow{2}[1]{*}{HAKE} & \multirow{2}[1]{*}{Base} & \multirow{2}[1]{*}{0.5} & -- \\
                         &       & MIX   & $^{\dagger}${\textbf{50.2}} & 0.2   & $^{\dagger}${\textbf{41.0}} & 0.4   & $^{\dagger}${\textbf{55.8}} & 0.1   & \textbf{67.5} & 0.2   &       &       &       & 0.5 \\
\cmidrule{2-15}          & \multirow{1.5}[6]{*}{Uniq} & CBS   & \textbf{49.8} & 0.2   & \textbf{40.3} & 0.2   & \textbf{55.4} & 0.1   & $^{\dagger}${\textbf{67.6}} & 0.1   &       &       &       &  \\
\cmidrule{3-15}          &       & MBS   & 49.5  & 0.2   & 39.9  & 0.2   & 55.2  & 0.3   & 67.4  & 0.2   & \multirow{2}[1]{*}{RotatE} & \multirow{2}[1]{*}{Base} & \multirow{2}[1]{*}{0.5} & -- \\
                         &       & MIX   & 49.7  & 0.2   & \textbf{40.3} & 0.2   & \textbf{55.4} & 0.2   & 67.5  & 0.2   &       &       &       & 0.5 \\
    \midrule
    \multirow{7.5}[20]{*}{HAKE} & \multicolumn{2}{c}{None} & 53.6  & 0.1   & 45.0  & 0.3   & 58.9  & 0.3   & 69.0  & 0.0   &       &       &       &  \\
\cmidrule{2-15}          & \multirow{1.5}[6]{*}{Base} & CBS   & \textbf{54.3} & 0.1   & \textbf{45.9} & 0.2   & \textbf{59.6} & 0.2   & \textbf{69.3} & 0.1   &       &       &       &  \\
\cmidrule{3-15}          &       & MBS   & 53.6  & 0.3   & 44.9  & 0.4   & 58.9  & 0.2   & 68.8  & 0.1   & \multirow{2}[1]{*}{HAKE} & \multirow{2}[1]{*}{None} & \multirow{2}[1]{*}{0.1} & -- \\
                         &       & MIX   & 54.0  & 0.1   & 45.4  & 0.1   & 59.3  & 0.3   & 69.2  & 0.1   &       &       &       & 0.5 \\
\cmidrule{2-15}          & \multirow{1.5}[6]{*}{Freq} & CBS   & 54.5  & 0.3   & 46.1  & 0.3   & 59.8  & 0.5   & 69.4  & 0.3   &       &       &       &  \\
\cmidrule{3-15}          &       & MBS   & \textbf{54.8} & 0.1   & 46.5  & 0.2   & \textbf{60.0} & 0.3   & \textbf{69.7} & 0.1   & \multirow{2}[1]{*}{RotatE} & \multirow{2}[1]{*}{None} & \multirow{2}[1]{*}{0.5} & -- \\
                         &       & MIX   & \textbf{54.8} & 0.1   & \textbf{46.7} & 0.1   & 59.7  & 0.2   & 69.5  & 0.1   &       &       &       & 0.1 \\
\cmidrule{2-15}          & \multirow{1.5}[6]{*}{Uniq} & CBS   & $^{\dagger}${\textbf{55.1}} & 0.1   & $^{\dagger}${\textbf{46.8}} & 0.2   & $^{\dagger}${\textbf{60.1}} & 0.3   & $^{\dagger}${\textbf{70.0}} & 0.2   &       &       &       &  \\
\cmidrule{3-15}          &       & MBS   & 54.8  & 0.1   & 46.5  & 0.2   & 60.0  & 0.3   & 69.7  & 0.1   & \multirow{2}[1]{*}{RotatE} & \multirow{2}[1]{*}{None} & \multirow{2}[1]{*}{0.5} & -- \\
                         &       & MIX   & 54.9  & 0.1   & 46.6  & 0.1   & 60.0  & 0.2   & 69.9  & 0.2   &       &       &       & 0.3 \\
    \bottomrule
    \end{tabular}}
    }%
    \caption{Results on YAGO3-10. The notations are the same as the ones in Table \ref{tab:fb15k-237}.\label{tab:YAGO3-10}}
\end{table*}%

\begin{figure*}[t]
    \centering
    \includegraphics[width=\textwidth]{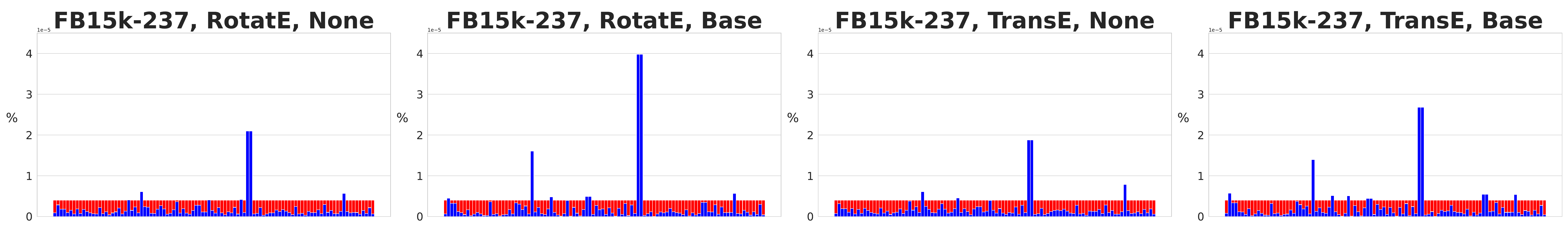}
    \includegraphics[width=\textwidth]{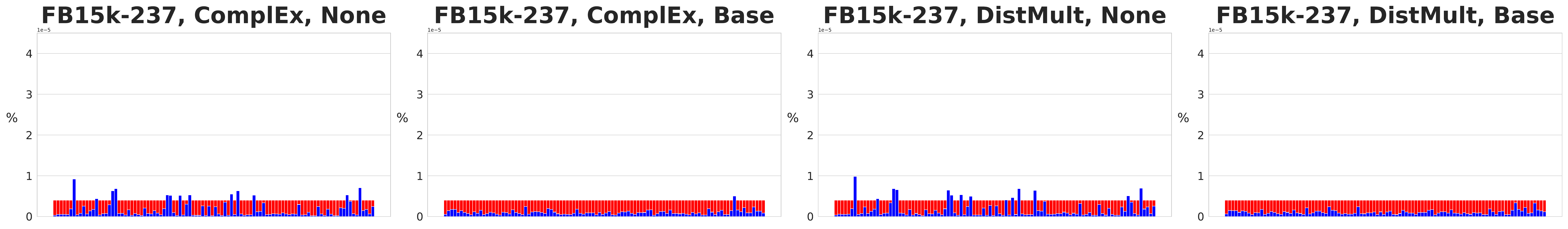}
    \includegraphics[width=\textwidth]{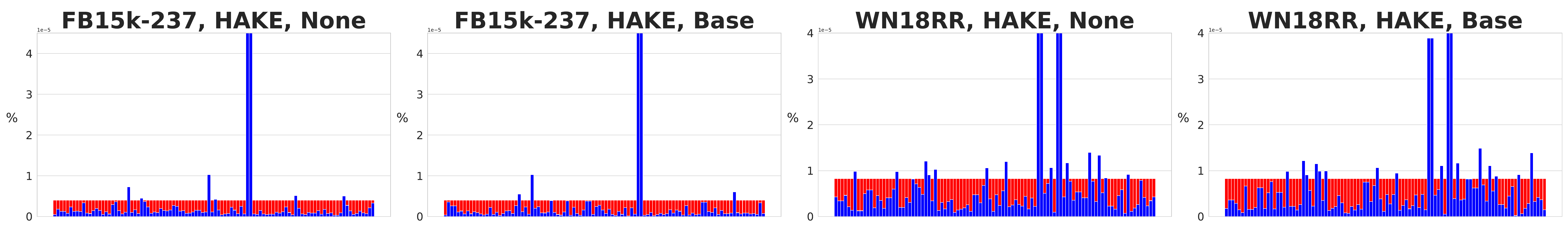}
    \includegraphics[width=\textwidth]{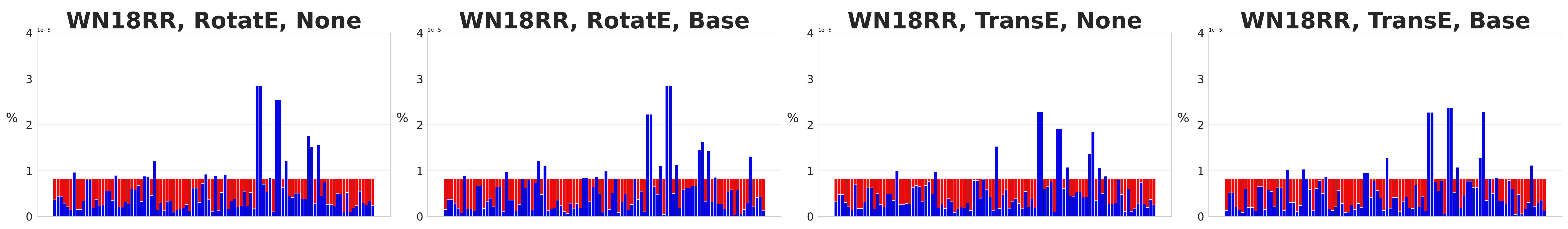}
    \includegraphics[width=\textwidth]{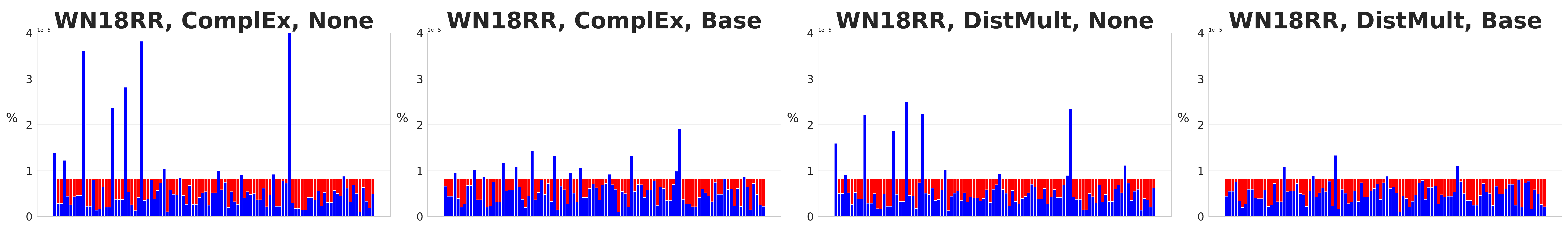}
    \includegraphics[width=\textwidth]{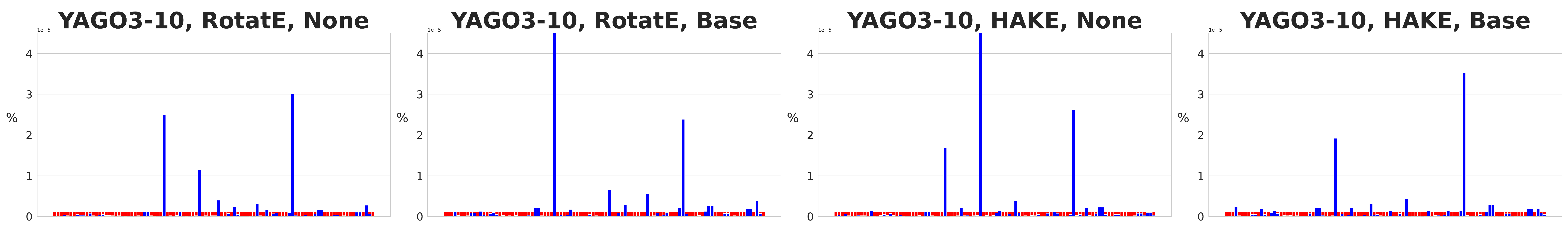}
    \includegraphics[width=\textwidth]{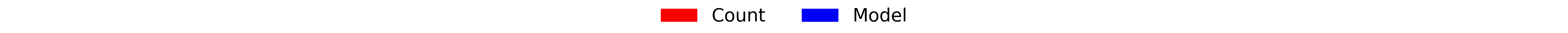}
    \caption{Appearance probabilities (\%) of queries in CBS and MBS that have the lowest 100 CBS frequencies for each setting, sorted left to right in descending order by their CBS frequencies.}
    \label{fig:normed_freq}
\end{figure*}

\paragraph{Metrics} 
We evaluated these methods using the most conventional metrics in KGC, i.e., Mean Reciprocal Rank (MRR), Hits@1 (H@1), Hits@3 (H@3), and Hits@10 (H@10). We reported the average scores in three different runs by changing their seeds\footnotemark~for each metric. We also reported the standard deviations of the scores by the three runs.

\footnotetext{We fixed seed numbers for the three trials in the training model and sub-model correspondingly. Note that the appearance probabilities drawn in Figure \ref{fig:normed_freq} all use the same seed.}

\paragraph{Implementations and Hyper-parameters}

For RotatE, TransE, ComplEx, and DistMult, we followed the implementations and hyper-parameters reported by \citet{rotate}. For HAKE, we inherited the setting of \citet{https://doi.org/10.48550/arxiv.1911.09419}.

In our experiments, the performance of subsampling is influenced by the selection of the following hyper-parameters: (1) temperature $\alpha$; (2) $\lambda$, the ratio of MBS against CBS.
For our proposed MBS subsampling, we chose $\alpha$ from $\{2.0,1.0,0.5,0.1,0.05,0.01\}$ based on validation MRR. 
For our proposed MIX subsampling, we inherited the best $\alpha$ in MBS. Then, we chose the mix ratio $\lambda$ from $\{0.1,0.3,0.5,0.7,0.9\}$ based on validation MRR.

In FB15k-237 and WN18RR, we chose the sub-model from RotatE, TransE, HAKE, ComplEx, and DistMult with the setting of Base and None based on the validation MRR.
In YAGO3-10, we also chose the sub-model from RotatE and HAKE, similar to FB15k-237 and WN18RR.

\subsection{Results}

\paragraph{Results} 
Table \ref{tab:fb15k-237}, \ref{tab:wn18rr}, and \ref{tab:YAGO3-10} show the KGC performances on FB15k-237, WN18RR and YAGO3-10, respectively. 
Note that the results of Wilcoxon signed-rank test for performance differences between MBS/MIX and CBS show statistical significance with p-values less than 0.01 in all cases when MBS/MIX outperforms CBS.

As we can see, the models trained with MIX or MBS achieved the best results in all models on FB15k-237 and WN18RR. However, in YAGO3-10, HAKE with Freq in CBS outperformed the results of MBS and MIX. Considering that the preprocess of YAGO3-10 filtered out entities with less than 10 relations in the dataset, we can conclude that MBS and MIX are effective on the sparse KGs like that of FB15k-237 and WN18RR. These results are along with our expectation that MBS and MIX can improve the completion performances in sparse KGs as introduced in \S\ref{sec:intro}.

In individual comparison for each metric, CBS sometimes outperformed MIX or MBS. This is because the estimated frequencies in MIX and MBS rely on selected sub-models.
From these results, we can understand that MIX and MBS have the potential to improve the KG completion performances by carefully choosing their sub-model.

\paragraph{Analysis} 

We analyze the remaining question, i.e., which sub-model to choose for MBS. Table \ref{tab:fb15k-237}, \ref{tab:wn18rr}, and \ref{tab:YAGO3-10} show the selected sub-models for each MBS (See \S\ref{sec:setting} in details), where ComplEx dominates over other models in FB15k-237 and WN18RR. To know the reason, we depict MBS frequencies of queries that have the bottom 100 CBS frequencies in Figure \ref{fig:normed_freq}. 
In FB15k-237, we can see several spikes of frequencies in TransE, RotatE, and HAKE that do not exist in ComplEx. 
In WN18RR, the peak frequencies of ComplEx with None are larger and broader than that of other sub-models. 
These results indicate that models in FB15k-237 and WN18RR, respectively, encountered problems of an over and lack of smoothing, and MBS dealt with this problem.
Because sparseness is a problem when data is small, these are along with the fact that FB15k-237 has larger training data than WN18RR. Thus, choosing a suitable sub-model for a target dataset is important in MBS.

\paragraph{Discussion}

We discuss how sub-model and hyper-parameter choices contribute to the improvement of KGE performance apart from our method.  
The choice of the sub-model and the $\alpha$ played significant roles in the observed improvements because distributions from sub-model prediction depend on each sub-model and each dataset. Since we adopted the value of $\alpha$ used in the past state-of-the-art method of \citet{rotate} and \citet{https://doi.org/10.48550/arxiv.1911.09419}, we believe that the performance gains of MBS are not only caused by the values of $\alpha$. Similarly, keeping $\lambda$ constant in the MIX strategy may lead to certain improvements depending on used sub-models and datasets. However, as shown in Appendix \ref{appendix:mix-multi-task}, $\lambda$ has the role of adjusting the loss of multi-task learning, and thus, it may be more sensitive compared with $\alpha$. 

\section{Related Work}

\citet{ns} originally propose the NS loss to train their word embedding model, word2vec. \citet{Tro16} introduce the NS loss to KGE to reduce training time. \citet{rotate} extend the NS loss for KGE by introducing a margin term, normalization of negative samples, and newly proposed their noise distribution. \citet{kamigaito-hayashi-2021-unified} claim the importance of dealing with the sparseness problem of KGs through their theoretical analysis of the NS loss in KGE. Furthermore, \citet{pmlr-v162-kamigaito22a} reveal that subsampling \cite{ns} can alleviate the sparseness problem in the NS for KGE.

Similar to these works, our work aims to investigate and extend the NS loss used in KGE to improve KG performance.

\section{Conclusion}

In this paper, we propose new subsampling approaches, MBS and MIX, that can deal with the problem of low-frequent entity-relation pairs in CBS by estimating their frequencies using the sub-model prediction.
Evaluation results on FB15k-237 and WN18RR showed the improvement of KGC performances by MBS and MIX.
Furthermore, our analysis also revealed that selecting an appropriate sub-model for the target dataset is important for improving KGC performances.

\section*{Limitations}

Utilizing our model-based subsampling requires pre-training for choosing a suitable sub-model, and thus may require more than twice the computational budget. However, since we can use a small model as a sub-model, like the use of ComplEx as a sub-model for HAKE, there is a possibility that the actual computational cost becomes less than the doubled one.

For calculating CBS frequencies, we only use the one with the arithmetic mean since we inherited the conventional subsampling methods as our baseline. Thus, we can consider various replacements not covered by this paper for the operation. However, even if we carefully choose the operation, CBS is essentially difficult to induce the appropriate appearance probabilities of low-frequent queries compared with our MBS, which can use vector-space embedding.

Our experiments are carried out only on FB15k-237, WN18RR, and YAGO3-10 datasets. Thus, whether our method works for larger and noisier data is to be verified.

Furthermore, although our method is generalizable to deep learning models, our current work is conducted purely on KGE models, and whether it works for general deep learning models as well is to be verified.

\section*{Acknowledgements}
This work was supported by NAIST Touch Stone, i.e., JST SPRING Grant Number JPMJSP2140, and JSPS KAKENHI Grant Numbers JP21H05054 and JP23H03458.

\bibliography{anthology,custom}

\begin{thebibliography}{16}
\expandafter\ifx\csname natexlab\endcsname\relax\def\natexlab#1{#1}\fi

\bibitem[{Bordes et~al.(2013)Bordes, Usunier, Garcia-Duran, Weston, and
  Yakhnenko}]{Bor13}
Antoine Bordes, Nicolas Usunier, Alberto Garcia-Duran, Jason Weston, and Oksana
  Yakhnenko. 2013.
\newblock Translating embeddings for modeling multi-relational data.
\newblock In \emph{Advances in Neural Information Processing Systems},
  volume~26. Curran Associates, Inc.

\bibitem[{Dettmers et~al.(2018)Dettmers, Minervini, Stenetorp, and
  Riedel}]{Det18}
Tim Dettmers, Pasquale Minervini, Pontus Stenetorp, and Sebastian Riedel. 2018.
\newblock Convolutional 2d knowledge graph embeddings.
\newblock In \emph{AAAI}.

\bibitem[{Guan et~al.(2019)Guan, Wang, and Huang}]{Guan_Wang_Huang_2019}
Jian Guan, Yansen Wang, and Minlie Huang. 2019.
\newblock \href {https://doi.org/10.1609/aaai.v33i01.33016473} {Story ending
  generation with incremental encoding and commonsense knowledge}.
\newblock \emph{Proceedings of the AAAI Conference on Artificial Intelligence},
  33(01):6473--6480.

\bibitem[{Kamigaito and Hayashi(2021)}]{kamigaito-hayashi-2021-unified}
Hidetaka Kamigaito and Katsuhiko Hayashi. 2021.
\newblock \href {https://doi.org/10.18653/v1/2021.acl-long.429} {Unified
  interpretation of softmax cross-entropy and negative sampling: With case
  study for knowledge graph embedding}.
\newblock In \emph{Proceedings of the 59th Annual Meeting of the Association
  for Computational Linguistics and the 11th International Joint Conference on
  Natural Language Processing (Volume 1: Long Papers)}, pages 5517--5531,
  Online. Association for Computational Linguistics.

\bibitem[{Kamigaito and Hayashi(2022{\natexlab{a}})}]{pmlr-v162-kamigaito22a}
Hidetaka Kamigaito and Katsuhiko Hayashi. 2022{\natexlab{a}}.
\newblock \href {https://proceedings.mlr.press/v162/kamigaito22a.html}
  {Comprehensive analysis of negative sampling in knowledge graph
  representation learning}.
\newblock In \emph{Proceedings of the 39th International Conference on Machine
  Learning}, volume 162 of \emph{Proceedings of Machine Learning Research},
  pages 10661--10675. PMLR.

\bibitem[{Kamigaito and Hayashi(2022{\natexlab{b}})}]{kamigaito2022subsampling}
Hidetaka Kamigaito and Katsuhiko Hayashi. 2022{\natexlab{b}}.
\newblock \href {http://arxiv.org/abs/2209.12801} {Subsampling for knowledge
  graph embedding explained}.

\bibitem[{Katz(1987)}]{katz1987estimation}
Slava Katz. 1987.
\newblock Estimation of probabilities from sparse data for the language model
  component of a speech recognizer.
\newblock \emph{IEEE transactions on acoustics, speech, and signal processing},
  35(3):400--401.

\bibitem[{Lukovnikov et~al.(2017)Lukovnikov, Fischer, Lehmann, and
  Auer}]{10.1145/3038912.3052675}
Denis Lukovnikov, Asja Fischer, Jens Lehmann, and S\"{o}ren Auer. 2017.
\newblock \href {https://doi.org/10.1145/3038912.3052675} {Neural network-based
  question answering over knowledge graphs on word and character level}.
\newblock In \emph{Proceedings of the 26th International Conference on World
  Wide Web}, WWW '17, page 1211–1220, Republic and Canton of Geneva, CHE.
  International World Wide Web Conferences Steering Committee.

\bibitem[{Mikolov et~al.(2013)Mikolov, Sutskever, Chen, Corrado, and Dean}]{ns}
Tom{\'{a}}s Mikolov, Ilya Sutskever, Kai Chen, Greg Corrado, and Jeffrey Dean.
  2013.
\newblock \href {http://arxiv.org/abs/1310.4546} {Distributed representations
  of words and phrases and their compositionality}.
\newblock \emph{CoRR}, abs/1310.4546.

\bibitem[{Moon et~al.(2019)Moon, Shah, Kumar, and
  Subba}]{moon-etal-2019-opendialkg}
Seungwhan Moon, Pararth Shah, Anuj Kumar, and Rajen Subba. 2019.
\newblock \href {https://doi.org/10.18653/v1/P19-1081} {{O}pen{D}ial{KG}:
  Explainable conversational reasoning with attention-based walks over
  knowledge graphs}.
\newblock In \emph{Proceedings of the 57th Annual Meeting of the Association
  for Computational Linguistics}, pages 845--854, Florence, Italy. Association
  for Computational Linguistics.

\bibitem[{Neubig and Dyer(2016)}]{neubig-dyer-2016-generalizing}
Graham Neubig and Chris Dyer. 2016.
\newblock \href {https://doi.org/10.18653/v1/D16-1124} {Generalizing and
  hybridizing count-based and neural language models}.
\newblock In \emph{Proceedings of the 2016 Conference on Empirical Methods in
  Natural Language Processing}, pages 1163--1172, Austin, Texas. Association
  for Computational Linguistics.

\bibitem[{Sun et~al.(2019)Sun, Deng, Nie, and Tang}]{rotate}
Zhiqing Sun, Zhi{-}Hong Deng, Jian{-}Yun Nie, and Jian Tang. 2019.
\newblock \href {https://openreview.net/forum?id=HkgEQnRqYQ} {Rotate: Knowledge
  graph embedding by relational rotation in complex space}.
\newblock In \emph{Proceedings of the 7th International Conference on Learning
  Representations, {ICLR} 2019}.

\bibitem[{Toutanova and Chen(2015)}]{Tou15}
Kristina Toutanova and Danqi Chen. 2015.
\newblock \href {https://doi.org/10.18653/v1/W15-4007} {Observed versus latent
  features for knowledge base and text inference}.
\newblock In \emph{Proceedings of the 3rd Workshop on Continuous Vector Space
  Models and their Compositionality}, pages 57--66, Beijing, China. Association
  for Computational Linguistics.

\bibitem[{Trouillon et~al.(2016)Trouillon, Welbl, Riedel, Gaussier, and
  Bouchard}]{Tro16}
Th{\'{e}}o Trouillon, Johannes Welbl, Sebastian Riedel, {\'{E}}ric Gaussier,
  and Guillaume Bouchard. 2016.
\newblock \href {http://arxiv.org/abs/1606.06357} {Complex embeddings for
  simple link prediction}.
\newblock \emph{CoRR}, abs/1606.06357.

\bibitem[{Yang et~al.(2015)Yang, tau Yih, He, Gao, and
  Deng}]{yang2015embedding}
Bishan Yang, Wen tau Yih, Xiaodong He, Jianfeng Gao, and Li~Deng. 2015.
\newblock \href {http://arxiv.org/abs/1412.6575} {Embedding entities and
  relations for learning and inference in knowledge bases}.

\bibitem[{Zhang et~al.(2019)Zhang, Cai, Zhang, and
  Wang}]{https://doi.org/10.48550/arxiv.1911.09419}
Zhanqiu Zhang, Jianyu Cai, Yongdong Zhang, and Jie Wang. 2019.
\newblock \href {https://doi.org/10.48550/ARXIV.1911.09419} {Learning
  hierarchy-aware knowledge graph embeddings for link prediction}.

\end{thebibliography}
\bibliographystyle{acl_natbib}

\clearpage
\appendix
\onecolumn

\section{Note on Figure \ref{fig:effect_subsampling}}
\label{appendix:settings}
To illustrate the results on FB15k-237 and WN18RR datasets, we used TransE, RotatE, ComplEx, DistMult, and HAKE as the KGE models. To plot that on the YAGO3-10 dataset, we used RotatE and HAKE as the KGE models following the setting in \S\ref{sec:setting}.
Regarding the use of subsampling, the MRR scores of using subsampling refer to the result of Base subsampling in Table \ref{tab:subsampling} with CBS frequencies, whereas that without subsampling corresponds to the setting "None" \S\ref{sec:setting}.

\section{Interpretation of MIX Subsampling as Multi-task Learning}
\label{appendix:mix-multi-task}

We can reformulate Eq.~(\ref{eq:loss:mix}) as follows:
\begin{align}
    &\ell_{mix}(\mathbf{\theta};\mathbf{\theta}')\\
    =&-\frac{1}{|D|}\sum_{(x,y) \in D} \Bigl[A_{mix}(\mathbf{\theta}')\log(\sigma(s_{\mathbf{\theta}}(x,y)+\gamma))+\frac{1}{\nu}\!\!\!\!\!\!\sum_{y_{i}\sim p_n(y_{i}|x)}^{\nu}\!\!\!\!\!\!B_{mix}(\mathbf{\theta}')\log(\sigma(-s_{\mathbf{\theta}}(x,y_i)-\gamma))\Bigr],\\
    =&-\frac{1}{|D|}\sum_{(x,y) \in D} \Bigl[(\lambda A_{mbs}(\mathbf{\theta}') + (1-\lambda)A_{cbs})\log(\sigma(s_{\mathbf{\theta}}(x,y)+\gamma))\nonumber\\
    &+\frac{1}{\nu}\!\!\!\!\!\!\sum_{y_{i}\sim p_n(y_{i}|x)}^{\nu}\!\!\!\!\!\!(\lambda B_{mbs}(\mathbf{\theta}') + (1-\lambda)B_{cbs})\log(\sigma(-s_{\mathbf{\theta}}(x,y_i)-\gamma))\Bigr],\\
    =&-\frac{\lambda}{|D|}\sum_{(x,y) \in D} \Bigl[A_{mbs}(\mathbf{\theta}')\log(\sigma(s_{\mathbf{\theta}}(x,y)+\gamma))+\frac{1}{\nu}\!\!\!\!\!\!\sum_{y_{i}\sim p_n(y_{i}|x)}^{\nu}\!\!\!\!\!\!B_{mbs}(\mathbf{\theta}')\log(\sigma(-s_{\mathbf{\theta}}(x,y_i)-\gamma))\Bigr],\nonumber\\
    &-\frac{1-\lambda}{|D|}\sum_{(x,y) \in D} \Bigl[A_{cbs}\log(\sigma(s_{\mathbf{\theta}}(x,y)+\gamma))+\frac{1}{\nu}\!\!\!\!\!\!\sum_{y_{i}\sim p_n(y_{i}|x)}^{\nu}\!\!\!\!\!\!B_{cbs}\log(\sigma(-s_{\mathbf{\theta}}(x,y_i)-\gamma))\Bigr],\\
    =&\lambda\ell_{mbs}(\mathbf{\theta};\mathbf{\theta}') + (1-\lambda)\ell_{cbs}(\mathbf{\theta}) \label{eq:loss:multi}
\end{align}
From Eq.~(\ref{eq:loss:multi}), since $\ell_{mix}(\mathbf{\theta};\mathbf{\theta}')$ is the mixed loss of the two loss functions $\ell_{mbs}(\mathbf{\theta};\mathbf{\theta}')$ and $\ell_{cbs}(\mathbf{\theta})$, we can understand that using MIX is multi-task learning of using both CBS and MBS.

\end{document}